\documentclass[11pt]{article}

\usepackage{acl}
\usepackage{subcaption}
\usepackage{times}
\usepackage{latexsym}
\usepackage{balance}
\usepackage{amsmath}
\usepackage{amssymb}
\usepackage{booktabs}
\usepackage{graphicx}
\usepackage{booktabs}
\usepackage{colortbl}    % \rowcolor
\usepackage{xcolor}
\usepackage{multirow}

\usepackage[T1]{fontenc}
\usepackage[utf8]{inputenc}

\usepackage{microtype}

\usepackage{inconsolata}

\usepackage{graphicx}

\usepackage{xcolor}
\usepackage[skins,breakable]{tcolorbox}
\usepackage{amsmath}
\usepackage{listings}
\usepackage{caption}

\definecolor{solblue}{HTML}{A4C2F4}   % 标题栏浅蓝色
\definecolor{solbg}{HTML}{F4F8FC}     % 内容区极浅蓝/灰背景色
\definecolor{codebg}{HTML}{F8F9FA}    % 代码块背景色
\definecolor{errorred}{HTML}{FF0000}  % 错误标记的红色

\newtcolorbox{solutionbox}[1]{
    enhanced,
    breakable,                   % 允许跨页
    colback=solbg,               % 内容区背景色
    colframe=solblue,            % 边框颜色
    colbacktitle=solblue,        % 标题栏背景色
    coltitle=white,              % 标题文字颜色
    fonttitle=\bfseries\sffamily\large, % 标题字体（加粗、无衬线）
    title={#1},                  % 接收外部传入的标题（如 Solution 1）
    boxrule=1pt,                 % 边框粗细
    arc=4pt,                     % 圆角大小
    left=3mm, right=3mm,         % 左右内边距
    top=2mm, bottom=2mm,         % 上下内边距
    toptitle=1.5mm, bottomtitle=1.5mm, % 标题栏上下内边距
}

\lstdefinestyle{mypython}{
    language=Python,
    backgroundcolor=\color{codebg},
    basicstyle=\ttfamily\small,
    keywordstyle=\color{blue}\bfseries,
    commentstyle=\color{green!50!black},
    stringstyle=\color{red!60!black},
    showstringspaces=false,
    breaklines=true,
    frame=single,
    rulecolor=\color{gray!30},
}
\title{Detecting and Mitigating the Correct-Answer Extinction Window in Test-Time Reinforcement Learning with Majority Voting}

\author{
  Hongxiang Lin$^{*}$, 
  Zhirui Kuai$^{*}$, 
  Erpeng Xue$^{\dagger}$,
  Lei Wang
  \\
  Meituan\\
  \texttt{linhx0@hotmail.com} \quad \texttt{\{kuaizhirui, wanglei46, xueerpeng\}@meituan.com}
}

\begin{document}
\maketitle
  \renewcommand{\thefootnote}{\fnsymbol{footnote}}
  \footnotetext[1]{Equal contribution.}
  \footnotetext[2]{Corresponding author.}
  \renewcommand{\thefootnote}{\arabic{footnote}}
\begin{abstract}
Test-time reinforcement learning (TTRL) reports substantial accuracy gains on mathematical reasoning benchmarks using majority vote as a pseudo-label signal. We argue these gains are systematically misinterpreted: most reflect sharpening of already-solvable problems rather than genuine learning, while problems corrupted from correct to incorrect outnumber truly learned ones, and this damage is irreversible once majority vote locks onto a wrong answer.
Per-problem tracking reveals that correct-answer signals in low-ability problems are briefly active before being permanently suppressed, a phenomenon we term the \textit{Correct-Answer Extinction Window}, with Flip Rate (FR) as its leading indicator. We thus propose TTRL-Guard, a lightweight framework with three mechanisms targeting the extinction window: Flip-Rate-Aware Reward Scaling (FRS) down-weights at-risk updates as FR declines, Minority-Preserving Sampling (MPS) retains gradient signal from minority correct answers, and Risk-Conditioned Sparse Updatings (RCSU) suspends updates on polarized problems. 
Experiments across three models and four benchmarks show that TTRL-Guard achieves the best average pass@1 on Qwen2.5-7B-Instruct and Qwen3-4B, improves relatively over TTRL by +54\% on AIME 2025. 
\footnote{Our code and implementation details are available at \url{https://github.com/linhxkkkk/TTRL-Guard}.}
\end{abstract}

\begin{figure}[t]
    \centering
    \begin{subfigure}{0.48\linewidth}
        \includegraphics[width=\linewidth]{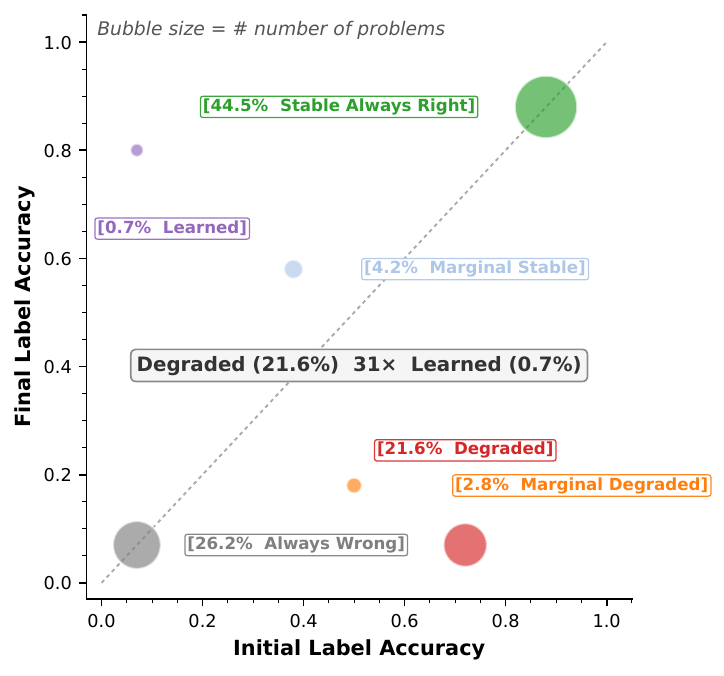}
        \caption{Label Migration}
        \label{fig:first}
    \end{subfigure}
    \hfill
    \begin{subfigure}{0.48\linewidth}
        \includegraphics[width=\linewidth]{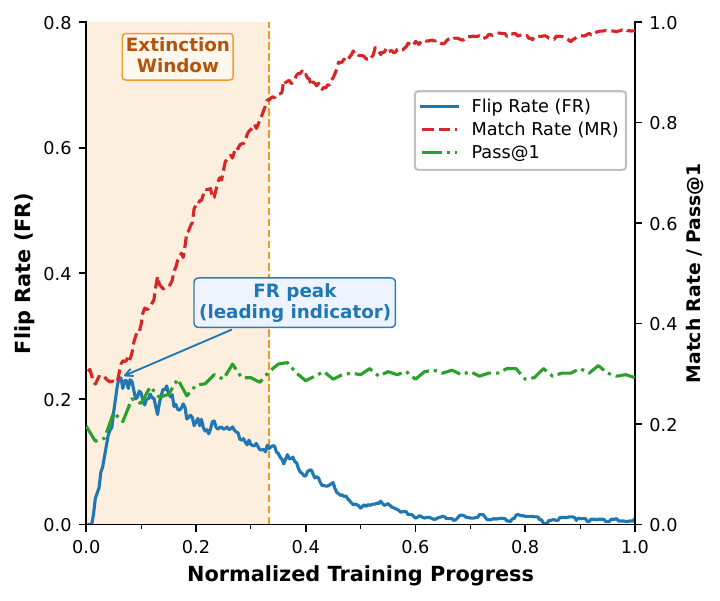}
        \caption{Training Dynamics}
        \label{fig:second}
    \end{subfigure}
    \caption{Label Dynamics During Test-Time Reinforcement Learning. (a) Label Migration Patterns Across Training. (b) Training Dynamics: Flip Rate as a Leading Indicator.}
    \label{fig:main}
\end{figure}

\section{Introduction}

Reinforcement learning with verifiable rewards (RLVR) has powered recent breakthroughs in LLM reasoning \citep{guo2025deepseek, comanici2025gemini, yang2025qwen3}. However, its reliance on ground-truth labels poses a fundamental scalability ceiling as tasks grow in complexity. This has spurred growing interest in unsupervised alternatives, most notably Test-time reinforcement learning (TTRL) \citep{Zuo2025TTRL}, which uses majority voting (MV) over sampled rollouts as pseudo-labels to enable self-improvement without human annotation. Alongside training-free distribution sharpening methods \citep{Ji2026Scalable}, unsupervised RL \cite{zhao2025learning, simonds2025ladder} is widely celebrated for pushing capability boundaries. Yet a critical question remains unanswered: \textit{Are these models truly learning to reason, or merely learning to agree with themselves?}

To answer this, we conduct a per-problem trajectory analysis across three models (Llama-3.2-3B-Instruct\cite{grattafiori2024llama}, Qwen2.5-7B-Instruct~\cite{yang2025qwen2.5}, Qwen3-4B \cite{yang2025qwen3}) and two reasoning benchmarks (AIME 2025~\cite{li2024numinamath}, MATH-500~\cite{hendrycks2021measuring}), comparing each problem's Initial Label Accuracy (ILA) and Final Label Accuracy (FLA) to classify problems into six categories (Details in Appendix~\ref{app:problem_categories}). As shown in Figure~\ref{fig:first}, 44.5\% of problems are already solvable before training (Stable Always Right); TTRL merely sharpens their pass@k toward 1.0, consistent with recent theoretical analyses \citep{he2026far}. Genuine zero-to-one capability acquisition accounts for a negligible 0.7\% (Learned). Meanwhile, 21.6\% of initially solvable problems suffer accuracy drops during training (Degraded), suggesting that optimization pressure on hard problems actively erodes performance on easier ones. The headline accuracy gains are thus largely an illusion of learning.

More alarmingly, this illusion masks a hidden cost. We term this \textit{Asymmetric Degradation}: problems corrupted from correct to incorrect outnumber genuinely learned ones by 31$\times$ (21.6\% vs.\ 0.7\%), yet degradation stays invisible in aggregate validation curves because it is dwarfed by the sharpening of already-solved problems. To understand this mechanism, we trace pseudo-label dynamics via two signals: the Flip Rate (FR), the fraction of problems whose majority-vote answer changes between consecutive steps, and the Match Rate (MR), the fraction of problems where the majority-vote answer matches the ground-truth label. As shown in Figure~\ref{fig:second} for Llama-3.2-3B-Instruct on AMC~\cite{li2024numinamath}, FR rises sharply in early training while correct answers still win the majority vote on over 31\% of problems. As false consensus solidifies, FR collapses and MR locks in the wrong answer, causing pass@1 on the test set to drop irreversibly. We formalize this brief window as the \textit{Correct-Answer Extinction Window}: once FR decays past a critical threshold, the pseudo-label cannot be recovered.

Recent work addresses majority-vote (MV) failures through external tool verification~\citep{liao2026tool}, step-wise confidence weighting~\citep{wang2025beyond}, difficulty-aware curricula~\citep{moradi2026disctt,yang2026ttcs}, and generator-verifier co-evolution~\citep{Pan2026CoVerRL}. Despite their diversity, these methods share a common limitation: they treat MV failure as a static property of the input and apply fixed interventions. To intercept this failure without external supervision, we propose \textbf{TTRL-Guard}, which monitors the FR as an online, label-free uncertainty signal and intervenes within the Extinction Window via three synergistic mechanisms: Flip-Rate-Aware Reward Scaling (FRS) down-weights updates on unstable pseudo-labels, Minority-Preserving Sampling (MPS) prolongs the survival of correct-minority signals, and Risk-Conditioned Sparse Updating (RCSU) halts updates once false consensus is formed. Experiments across three models and four benchmarks show that TTRL-Guard achieves the best average pass@1 on both Qwen2.5-7B-Instruct and Qwen3-4B, improves relatively over TTRL by +54\% on AIME 2025, and cuts the degraded-problem fraction from 60.2\% to 28.0\% on Llama-3.2-3B-Instruct.

Collectively, these findings reframe TTRL not as a general capability booster but as a sharpening mechanism bounded by the initial label accuracy distribution. Macro-level gains mask a fragile interior: pre-solvable problems benefit while the rest silently degrade, and the window for intervention is narrow. We find this window is most actionable in the moderate capability-difficulty regime (30-70\% initial pass@1); in extreme scenarios with near-zero or near-perfect initial accuracy, the extinction window barely opens. TTRL-Guard demonstrates that this window, once identified, can be acted upon without any external supervision, pointing toward a broader principle: reliable unsupervised RL requires not just better pseudo-labels, but dynamic awareness of when those labels can no longer be trusted.

\begin{figure*}[t]
    \centering

    % 第一张图
    \includegraphics[width=0.8\textwidth]{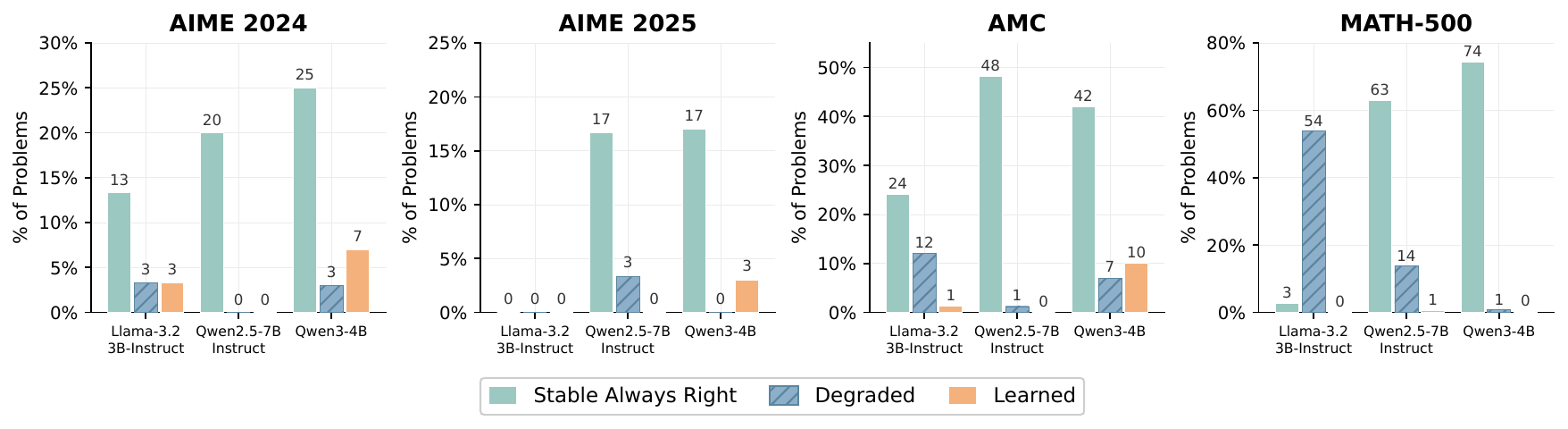}
    \caption{Problem-fate breakdown (Stable AR, Degraded, and Learned) for three models across four benchmarks after TTRL training.}
    \label{fig:cumulative_degraded}
    
    % 第二张图
    \includegraphics[width=0.8\textwidth]{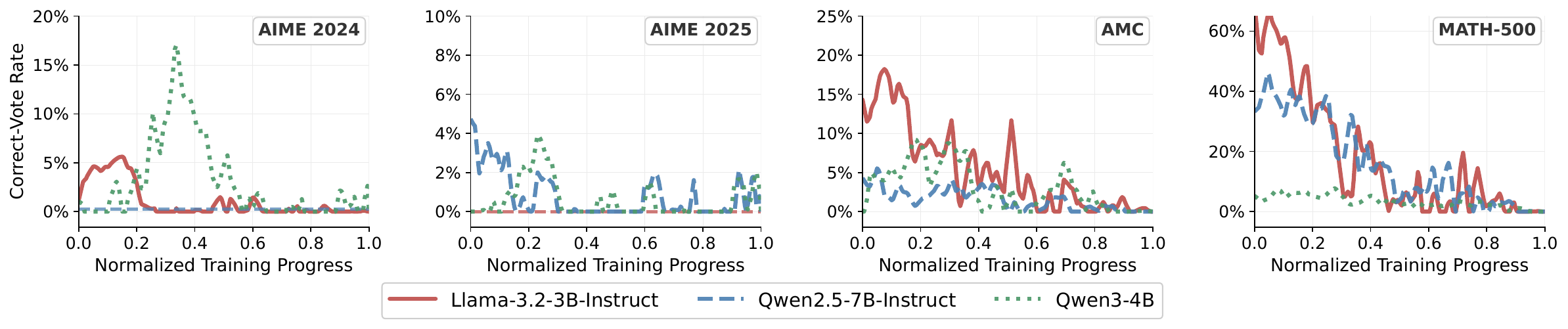}
    \caption{Correct-vote rate over normalized training progress, computed on the Degraded and Always-Wrong problem subsets.}
    \label{fig:la1_rate}

    % 第三张图
    \includegraphics[width=0.8\textwidth]{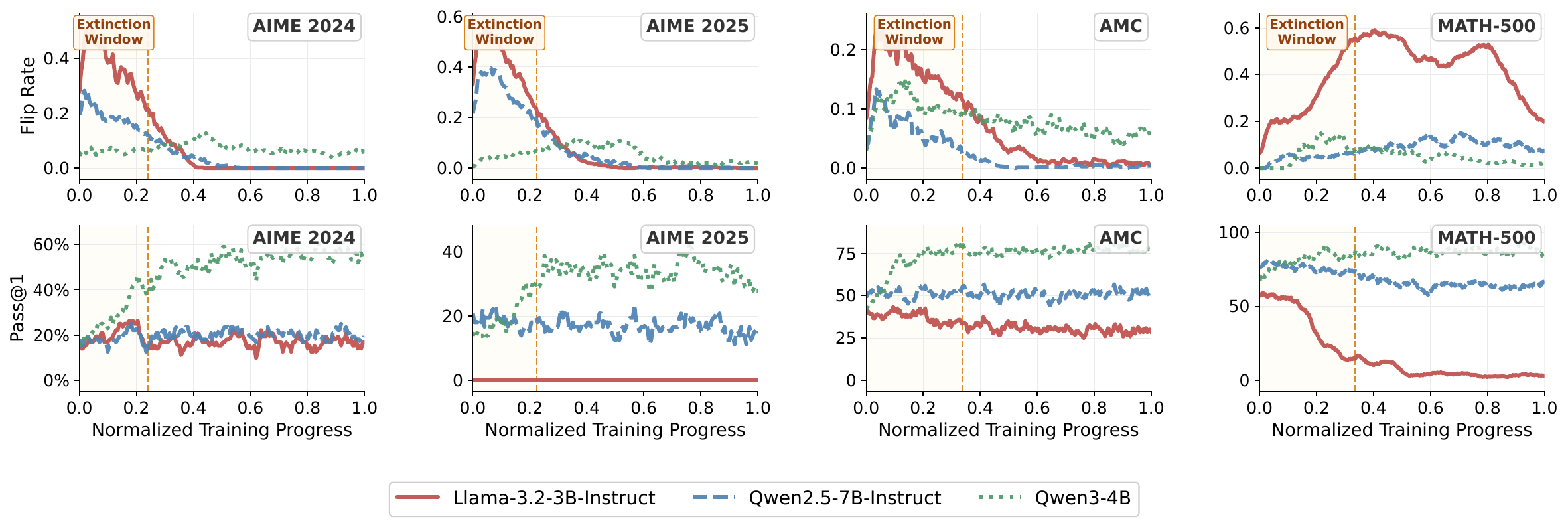}
    \caption{Flip Rate and Pass@1 over normalized training progress for each model across four benchmarks.}
    \label{fig:fr_la_trajectory}

\end{figure*}

\section{Dissecting TTRL Dynamics}
\label{sec:dissection}

We analyze TTRL training on three models (Llama-3.2-3B-Instruct, Qwen2.5-7B-Instruct, Qwen3-4B) and four datasets (AIME 2024, AIME 2025, AMC, MATH-500). Per-batch metrics include Match Rate (MR), the fraction of sampled responses matching the majority-vote winner; Flip Rate (FR), the fraction of problems whose majority-vote answer changes between consecutive steps; and Label Accuracy (LA), the per-problem fraction of responses matching the ground truth. Pass@1 is evaluated on the full test set periodically and serves as the primary aggregate metric.

\subsection{Problem Outcome Distribution}
\label{subsec:outcome_distribution}

We classify each problem into six outcomes: Stable Always Right (AR), Learned, Degraded, Marginal Stable, Marginal Degraded, and Always Wrong (Details in Appendix~\ref{app:problem_categories}). As shown in Figure~\ref{fig:cumulative_degraded}, Degraded problems far outnumber Learned problems by the end of training across most model-benchmark combinations. On MATH-500, Llama-3.2-3B-Instruct reaches 53.8\% Degraded versus 0.2\% Learned; Qwen2.5-7B-Instruct 13.8\% versus 0.6\%; Qwen3-4B 0.8\% versus 0.4\%. 

This asymmetry is capacity-dependent: Qwen3-4B is largely immune because 74.4\% of its problems are already Stable AR at initialization, so MV reinforces correct answers. Critically, this damage is invisible in Pass@1. Llama-3.2-3B-Instruct's Pass@1 appears stable or even improves slightly, even as 54\% of its MATH-500 problems suffer irreversible accuracy drops, demonstrating that aggregate accuracy curves mask individual problem degradation.

\subsection{The Correct-Answer Extinction Window}
\label{subsec:extinction_window}

For ultimately Degraded or Always Wrong problems, Figure~\ref{fig:la1_rate} plots the correct-vote rate, the fraction of steps where the correct answer wins majority voting, over normalized training progress. For Llama-3.2-3B-Instruct on MATH-500, this rate starts near 58\% but collapses to near zero by the final third of training. Once MV consolidates on the wrong answer, the correct gradient signal is permanently lost. The collapse arrives faster on harder datasets and is less pronounced in stronger models, confirming that the phenomenon is concentrated in weaker models on harder problems.

This irreversibility is a critical observation: we find that problems entering the Degraded category at any point rarely recover to Stable AR status in subsequent training, indicating that once false consensus forms, escape becomes increasingly unlikely under TTRL. This defines the \textit{Correct-Answer Extinction Window}: the brief early-training interval during which correct answers remain viable as minority signals before being permanently suppressed.

\subsection{Flip Rate as a Leading Indicator}
\label{subsec:flip_rate_indicator}

FR is computed entirely from model samples, providing a label-free early-warning signal for the extinction window. As shown in Figure~\ref{fig:fr_la_trajectory}, FR peaks when MV is still unstable, then collapses as consensus forms, before Pass@1 visibly drops. On MATH-500, where MV consensus forms more gradually, FR rises throughout the early phase before subsiding. Critically, during the high-FR phase, Pass@1 curves across models remain bundled; divergence appears only after FR has collapsed. This temporal ordering suggests that FR dynamics track the underlying label stability.

On MATH-500, the extinction window spans the first third of training; on AIME 2024/2025, it closes earlier due to higher problem difficulty. These observations motivate TTRL-Guard: the FR signal provides a real-time, label-free proxy for when the extinction window is open, enabling targeted intervention before correct-minority signals are permanently suppressed.

\begin{figure*}[t]
    \centering
    \includegraphics[width=0.8\textwidth]{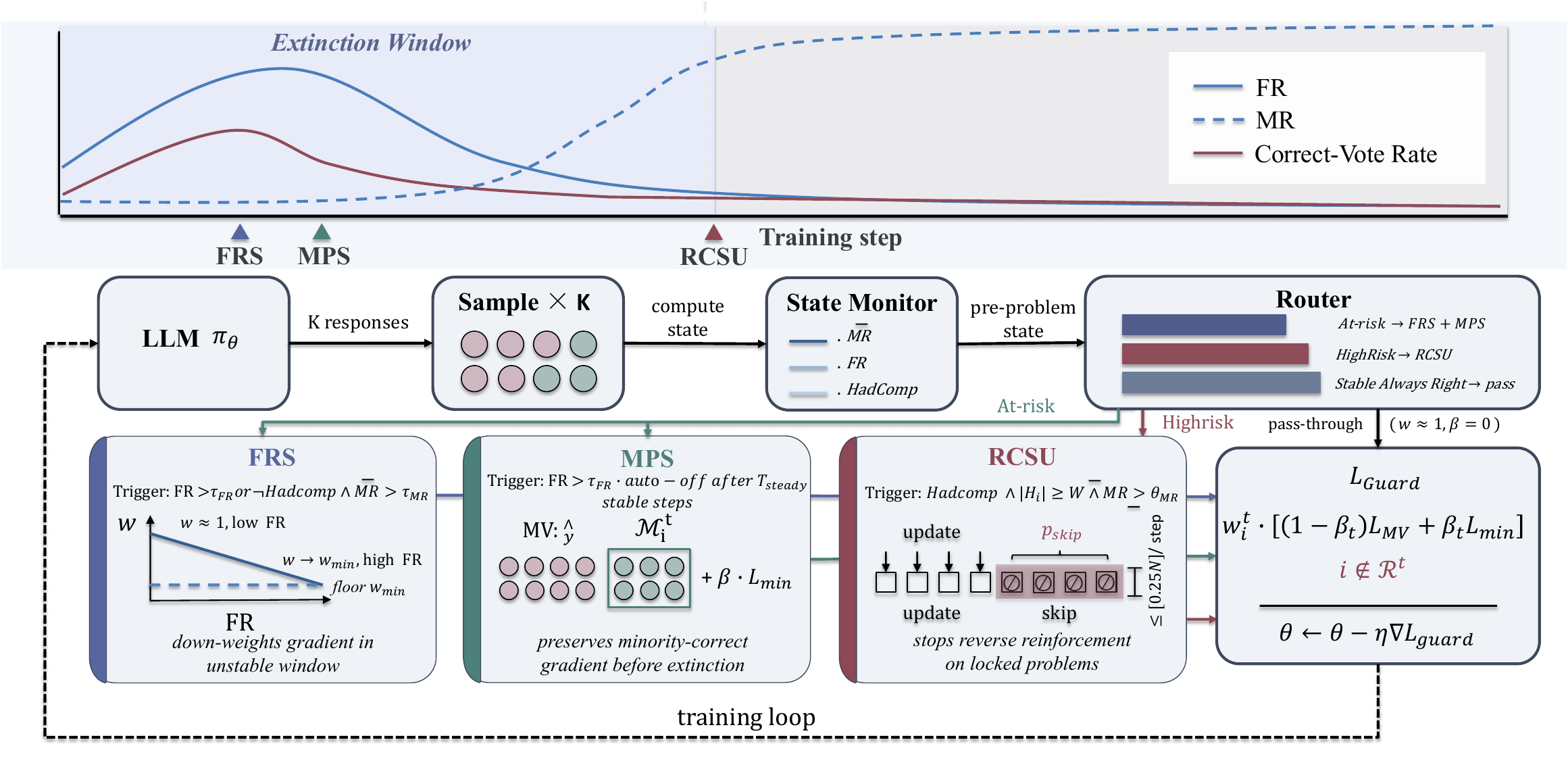}
    \caption{Overview of TTRL-Guard. The top panel illustrates the \textit{Extinction Window}, where the FR peaks early and decays while the correct-vote rate collapses irreversibly. The middle panel depicts the core pipeline for real-time problem monitoring and routing. The bottom panel details the three intervention pathways: at-risk problems enter FRS and MPS concurrently, high-risk cases are routed to RCSU, and stable problems bypass all modules unaltered.}
    % The top panel illustrates the correct-answer extinction window observed during TTRL training. The middle panel shows the core pipeline: the LLM samples $K$ responses per problem, a state monitor computes MR, FR, and HadComp, and a per-problem router dispatches each problem to one of three interventions. The bottom panel details FRS (flip-rate-aware reward scaling), MPS (minority-preserving sampling), and RCSU (risk-conditioned sparse updating), whose outputs are unified into $\mathcal{L}_{\text{Guard}}$ for gradient update.
    \label{fig:method}
\end{figure*}

\section{Method}

\label{sec:method}

The analysis in Section~\ref{sec:dissection} identifies the core failure mode of TTRL: during the early extinction window, MV systematically suppresses correct-minority signals, and once the wrong answer's match rate exceeds $1/2$, the pseudo-label becomes locked in an irreversible state. TTRL provides no mechanism to detect or counteract this process. We propose TTRL-Guard, a lightweight framework that intervenes precisely during this critical window by monitoring per-problem label dynamics in real time. As shown in Figure~\ref{fig:method}, TTRL-Guard acts only on at-risk problems before signals are permanently lost, leaving beneficial sharpening on mastered problems undisturbed and maintaining low computational overhead.

\subsection{Preliminary: TTRL with Majority-Voting}
\label{subsec:preliminary}

TTRL trains $\pi_\theta$ on a fixed unlabeled set $\mathcal{Q}=\{q_i\}_{i=1}^{N}$.
At each step $t$, the model draws $K$ responses $\{a_i^{(k)}\}_{k=1}^{K}$ per
problem and derives a pseudo-label via majority voting:
\begin{equation}
  \hat{y}_i^t = \arg\max_{a}\,\sum_{k=1}^{K}\mathbf{1}[a_i^{(k)}=a].
  \label{eq:mv}
\end{equation}
This pseudo-label serves as the reward target,
$r(a,\hat{y}_i^t)=\mathbf{1}[a=\hat{y}_i^t]$,
and the policy is optimized with GRPO~\cite{shao2024deepseekmath} to maximize
$\mathbb{E}_{a\sim\pi_\theta}[r(a,\hat{y}_i^t)]$.
As shown in Section~\ref{sec:dissection}, this reward is reliable only while
the correct answer holds a majority, a condition that collapses during the
extinction window. TTRL-Guard augments this loop with state monitoring per-problem and three targeted interventions.

\subsection{Per-Problem State Variables}
\label{subsec:formulation}

Building on Eq.~\eqref{eq:mv}, two additional step-level statistics are maintained:
\begin{align}
  \mathrm{MR}_i^t &= \frac{\max_a\sum_k\mathbf{1}[a_i^{(k)}=a]}{K},
  \label{eq:mr}\\
  \mathrm{FR}_i^t &= \frac{1}{W}\sum_{s=t-W+1}^{t}
                     \mathbf{1}[\hat{y}_i^s \neq \hat{y}_i^{s-1}],
  \label{eq:fr}
\end{align}
where $\mathrm{MR}_i^t$ (match rate) measures how dominant the leading answer is,
and $\mathrm{FR}_i^t$ (flip rate) is the fraction of the past $W$ steps in
which the pseudo-label changed, taking values in $[0,1]$.

Two cumulative state variables are additionally maintained without any
ground-truth labels:
\begin{align}
  \mathrm{HadComp}_i^t
    &= \bigvee_{s \leq t}\mathbf{1}[\mathrm{FR}_i^s > \tau_\mathrm{FR}],
    \label{eq:hadcomp}\\
  \bar{\mathrm{MR}}_i^{t}
    &= \frac{1}{\min(t,W)}\sum_{s=t-W+1}^{t}\mathrm{MR}_i^s,
    \label{eq:mrbar}
\end{align}
where $\tau_\mathrm{FR}$ is a flip-rate threshold and $W$ is a history window.
$\mathrm{HadComp}_i^t$ is set permanently once the windowed flip rate exceeds
$\tau_\mathrm{FR}$, distinguishing problems that have ever experienced
answer competition from those that have always been dominated by a single answer.
$\bar{\mathrm{MR}}_i^{t}$ is the sliding-mean match rate over the past $W$ steps.
As shown in Figure~\ref{fig:method} (top), declining FR foreshadows MV lock-in,
and $\mathrm{HadComp}$ separates benign consolidation (a Stable AR problem
settling on the correct answer) from dangerous consolidation (a wrong answer
taking over). These two variables are the sole observables driving all three
components.

\subsection{Flip-Rate-Aware Reward Scaling}
\label{subsec:frs}

As shown in Figure~\ref{fig:method} (top), pseudo-label reliability varies dramatically across the extinction window: during high-FR phases, the correct answer appears in 41.1\% of samples, but drops to 3.3\% once FR collapses (Llama-3.2-3B-Instruct on MATH-500, Section~\ref{subsec:extinction_window}). Uniform reward weights amplify wrong-answer gradients at these low-confidence moments. Flip-Rate-Aware Reward Scaling (FRS) down-weights rewards continuously with FR, creating adaptive credibility weights. Importantly, FRS scales only the reward weight, not the pseudo-label, preserving diverse response sampling.

The adjusted reward is
\begin{equation}
  \tilde{r}(a,\,\hat{y}_i^t) = r(a,\,\hat{y}_i^t)\cdot w_i^t,
  \label{eq:frs_reward}
\end{equation}
where $r(a,\hat{y}_i^t)=\mathbf{1}[a=\hat{y}_i^t]$ is the original binary
reward and the weight decomposes as $w_i^t = \alpha_i^t\cdot\gamma_i^t
\cdot\delta_i^t$, with
\begin{align}
  \alpha_i^t &= 1 - \lambda_1\cdot\mathrm{FR}_i^t,
  \label{eq:alpha}\\
  \gamma_i^t &= 1 - \lambda_2\cdot\mathbf{1}[C_1],
  \label{eq:gamma}\\
  \delta_i^t &= 1 - \tfrac{\lambda_2}{2}\cdot\mathbf{1}[C_2].
  \label{eq:delta}
\end{align}
The two trigger conditions are defined as
\begin{align}
  C_1 &:\;\mathrm{MR}_i^t > \tau_\mathrm{MR}
          \;\wedge\;\mathrm{FR}_i^t > \tau_\mathrm{FR},
  \label{eq:c1}\\
  C_2 &:\;\neg\mathrm{HadComp}_i
          \;\wedge\;|\mathcal{H}_i| \geq W
          \;\wedge\;\bar{\mathrm{MR}}_i^{t} > \tau_\mathrm{MR}.
  \label{eq:c2}
\end{align}
$C_1$ and $C_2$ are mutually exclusive by construction: $C_1$ requires active $\mathrm{HadComp}_i^t$, while $C_2$ requires it has never triggered, so at most one of $\gamma_i^t$ and $\delta_i^t$ deviates from 1 per step. $\alpha$ globally reduces reward magnitude during the unstable phase. $\gamma$ targets the contradictory state of high confidence with high flip rate ($C_1$), the signature of wrong-answer suppression (Section~\ref{subsec:flip_rate_indicator}). $\delta$ lightly penalizes never-challenged problems ($C_2$), likely stuck in wrong attractors; the trigger count is capped at $W$ to prevent gradient starvation. The weight is clipped at $w_i^t\geq w_{\min}$ to preserve non-zero gradients.

\subsection{Minority-Preserving Sampling}
\label{subsec:mps}

Section~\ref{subsec:extinction_window} shows that the correct answer survives as a competitive minority early in training before being systematically extinguished. Standard MV treats all non-majority responses as noise and discards their gradient signal entirely. This represents a loss of learnable information during the critical early window when the correct answer is still recoverable. Minority-Preserving Sampling (MPS) preserves this signal by assigning small positive rewards to minority answers that appear with sufficient frequency, effectively retaining them as weak learning targets. Unlike ~\cite{Pan2026CoVerRL} and T$^3$RL~\cite{Liao2026ToolVerification}, which require external verification or self-confidence scores, MPS operates entirely label-free, relying only on the relative frequency of responses in the sample.

MPS activates when $\mathrm{FR}_i^t > \tau_\mathrm{FR}$ and mixes the
standard MV objective with a minority-preservation term.
The minority candidate set is
\begin{equation}
  \mathcal{M}_i^t =
    \bigl\{a : \mathrm{vote}(a)\geq\lfloor K/4\rfloor,\;a\neq\hat{y}_i^t
    \bigr\},
  \label{eq:minority_set}
\end{equation}
where $\mathrm{vote}(a)$ counts responses whose extracted answer equals $a$.
Answers in $\mathcal{M}_i^t$ receive a small positive reward
$r_\mathrm{min}=\varepsilon\cdot\mathbf{1}[a\in\mathcal{M}_i^t]$,
where $\varepsilon\ll 1$ is a minority reward coefficient.
Here $\mathcal{L}_\mathrm{MV}(q_i)$ denotes the standard GRPO loss under the
majority-vote reward $r(a,\hat{y}_i^t)$ (Eq.~\eqref{eq:mv}), and
$\mathcal{L}_\mathrm{min}(q_i)$ is the minority-preservation loss computed
with reward $r_\mathrm{min}$ on answers in $\mathcal{M}_i^t$ and zero otherwise.
The training objective becomes
\begin{equation}
  \mathcal{L}_i^t =
    (1-\beta_t)\,\mathcal{L}_\mathrm{MV}(q_i)
    + \beta_t\,\mathcal{L}_\mathrm{min}(q_i),
  \label{eq:mps_obj}
\end{equation}
where $\beta_t\in[0,\beta_\mathrm{max}]$ scales linearly with
$\mathrm{FR}_i^t$.
Once $\mathrm{FR}_i^t$ drops below $\tau_\mathrm{FR}$, $\beta_t\to 0$ and
Eq.~\eqref{eq:mps_obj} recovers TTRL automatically.
If $\mathrm{FR}_i^t$ stays below $\tau_\mathrm{FR}$ for $T_\mathrm{steady}$
consecutive steps, minority protection deactivates completely, since a stably
dominant answer makes further diversity pressure counterproductive.

\subsection{Risk-Conditioned Sparse Updating}

\label{subsec:rcsu}

Once a wrong pseudo-label locks in, the extinction window closes and FR collapses. Continuing standard updates only accumulates reverse reinforcement. Risk-Conditioned Sparse Updating (RCSU) skips gradient updates on high-risk problems, halting harmful gradients while maintaining state monitoring.
A problem is flagged as high-risk when all three conditions hold: (i) $\mathrm{HadComp}_i^t$ (problem experienced answer competition), (ii) $|\mathcal{H}_i|\geq W$ (sufficient history accumulated), and (iii) $\bar{\mathrm{MR}}_i^{t}>\theta_\mathrm{MR}$ (high confidence in current answer). These identify problems once contested but now re-locked. Criterion (ii) uses history length rather than consecutive counts, since FR fluctuates on hard datasets causing resets before accumulating.
Problems are skipped with probability $p_\mathrm{skip}$ while maintaining state monitoring; stochastic skipping preserves recovery possibility if pseudo-labels flip. At most $\lfloor 0.25N\rfloor$ problems are skipped per step to maintain effective batch size.

\subsection{The TTRL-Guard Training Loop}
\label{subsec:training_loop}

Figure~\ref{fig:method} (middle and bottom) illustrates how the three
components integrate into a single training loop.
At each step, the model generates $K$ responses per problem and the State
Monitor updates $\hat{y}_i^t$, $\mathrm{MR}_i^t$, $\mathrm{FR}_i^t$,
$\mathrm{HadComp}_i^t$, and $\bar{\mathrm{MR}}_i^t$ for every problem.
The Router then assigns each problem to one of three paths: at-risk problems
($\mathrm{FR}_i^t>\tau_\mathrm{FR}$) enter FRS and MPS jointly; high-risk
problems ($\mathrm{HighRisk}_i^t=1$) are handled by RCSU; stable AR problems
bypass the intervention components with $w_i^t\approx 1$ and $\beta_t\approx 0$,
contributing directly to $\mathcal{L}_\mathrm{Guard}$ unchanged.

Formally, let $\mathcal{R}^t=\{i:\mathrm{HighRisk}_i^t=1\}$ denote the
high-risk set at step $t$.
The combined objective is
\begin{align}
  \mathcal{L}_\mathrm{Guard}
    &= \sum_{i\notin\mathcal{R}^t} w_i^t\cdot\mathcal{L}_i^t,
  \label{eq:guard}
\end{align}
where $w_i^t$ is the FRS weight from Eq.~\eqref{eq:frs_reward},
$\mathcal{L}_i^t$ is the per-problem objective from Eq.~\eqref{eq:mps_obj},
and problems in $\mathcal{R}^t$ are excluded with probability $p_\mathrm{skip}$.

\begin{table}[!t]
\centering
\resizebox{\columnwidth}{!}{%
\begin{tabular}{lcccccc}
\hline
\textbf{Method} &
  \textbf{AIME 2024} & \textbf{AIME 2025} &
  \textbf{AMC} & \textbf{MATH-500} & \textbf{Average} \\
\hline
\multicolumn{6}{c}{\textbf{\textit{Llama-3.2-3B-Instruct}}} \\
\hline
Base Model        &  4.7 &  0.0 & 20.6 & 43.9 & 17.3 \\
TTRL              & 13.3 &  0.0 & 31.3 & 52.1 & 24.1 \\
CoVerRL           & 16.9 &  \underline{1.8} & 34.8 & 57.0 & 27.6 \\
SCOPE             & 17.1 &  0.8 & \textbf{38.7} & \textbf{61.7} & \textbf{29.6} \\
\rowcolor{gray!10} 
TTRL (\textit{w/} FRS)        & 17.1 & \textbf{3.3} & 34.0 &54.0 & 27.1 \\
\rowcolor{gray!10} 
TTRL (\textit{w/} MPS)        & \textbf{20.3} & 0.8 & 33.3 & 55.8 & 27.6 \\
\rowcolor{gray!10} 
TTRL (\textit{w/} RCSU)        & 16.7 & 0.9 & 34.8 & 54.9 & 26.8 \\
\rowcolor{blue!10} 
\textbf{TTRL-Guard}       & \underline{19.5} & \textbf{3.3} & \underline{36.8} & \underline{57.3} & \underline{29.2} \\
\hline
\multicolumn{6}{c}{\textbf{\textit{Qwen2.5-7B-Instruct}}} \\
\hline
Base Model        & 18.1 &  6.4 & 43.6 & 77.2 & 36.3 \\
TTRL              & 20.0 & 15.6 & 52.4 & 81.0 & 42.3 \\
CoVerRL           & \underline{21.5} & 16.7 & \textbf{55.0} & 80.0 & 43.3 \\
SCOPE             & 21.1 & 14.6 & 51.1 & 81.0 & 42.0 \\
\rowcolor{gray!10}  
TTRL (\textit{w/} FRS)        & 20.2 & \underline{23.3} & 52.6 & \underline{82.1} & \underline{44.6} \\
\rowcolor{gray!10} 
TTRL (\textit{w/} MPS)        & 20.5 & 20.0 & 53.5 & \textbf{82.2} & 44.1 \\
\rowcolor{gray!10} 
TTRL (\textit{w/} RCSU)        & 21.3 & 20.1 & 51.2 & 81.6 & 43.6 \\
\rowcolor{blue!10} 
\textbf{TTRL-Guard}        & \textbf{21.9} & \textbf{24.1} & \underline{53.8} & 82.0 & \textbf{45.5} \\
\hline
\multicolumn{6}{c}{\textbf{\textit{Qwen3-4B}}} \\
\hline
Base Model        &  3.3 &  1.8 & 19.3 & 52.9 & 19.3 \\
TTRL              & 36.7 & 33.3 & 71.4 & 89.1 & 57.6 \\
CoVerRL           & \textbf{40.5} & 31.8 & 69.5 & 90.3 & 58.0 \\
SCOPE             & \underline{38.8} & 35.6 & 73.0 & 90.4 & \underline{59.5} \\
\rowcolor{gray!10} 
TTRL (\textit{w/} FRS)        & 35.5 & \underline{36.1} & 72.2 & 90.6 & 58.6 \\
\rowcolor{gray!10} 
TTRL (\textit{w/} MPS)        & 38.0 & 35.6 & \textbf{73.6} & \underline{90.7} & \underline{59.5} \\
\rowcolor{gray!10} 
TTRL (\textit{w/} RCSU)        & 36.4 & 35.9 & \underline{73.2} & 90.3 & 59.0 \\
\rowcolor{blue!10} 
\textbf{TTRL-Guard}        & 37.8 & \textbf{36.2} & \textbf{73.6} & \textbf{91.0} & \textbf{59.7} \\
\hline
\end{tabular}}
\caption{
Performance comparison of test-time adaptation methods (pass@1, \%).
Bold and underlined values indicate the best and second-best results, respectively.
}
\label{tab:main}
\end{table}
\section{Experiments}

% ------------------------------------------------------------------
\subsection{Experimental Setup}
% ------------------------------------------------------------------

\paragraph{Models and Datasets.} To evaluate the generalization ability of our method across diverse model architectures, we conduct experiments on three models spanning a wide range of parameter scales: Llama-3.2-3B-Instruct~\cite{grattafiori2024llama}, Qwen2.5-7B-Instruct~\cite{yang2025qwen2.5}, and Qwen3-4B~\cite{yang2025qwen3}. For training and evaluation, we employ four representative mathematical reasoning benchmarks: AIME 2024~\cite{li2024numinamath}, AIME 2025~\cite{li2024numinamath}, AMC~\cite{li2024numinamath}, and MATH-500~\cite{hendrycks2021measuring}.

\begin{figure*}[t]
  \centering
  \includegraphics[width=0.8\textwidth]{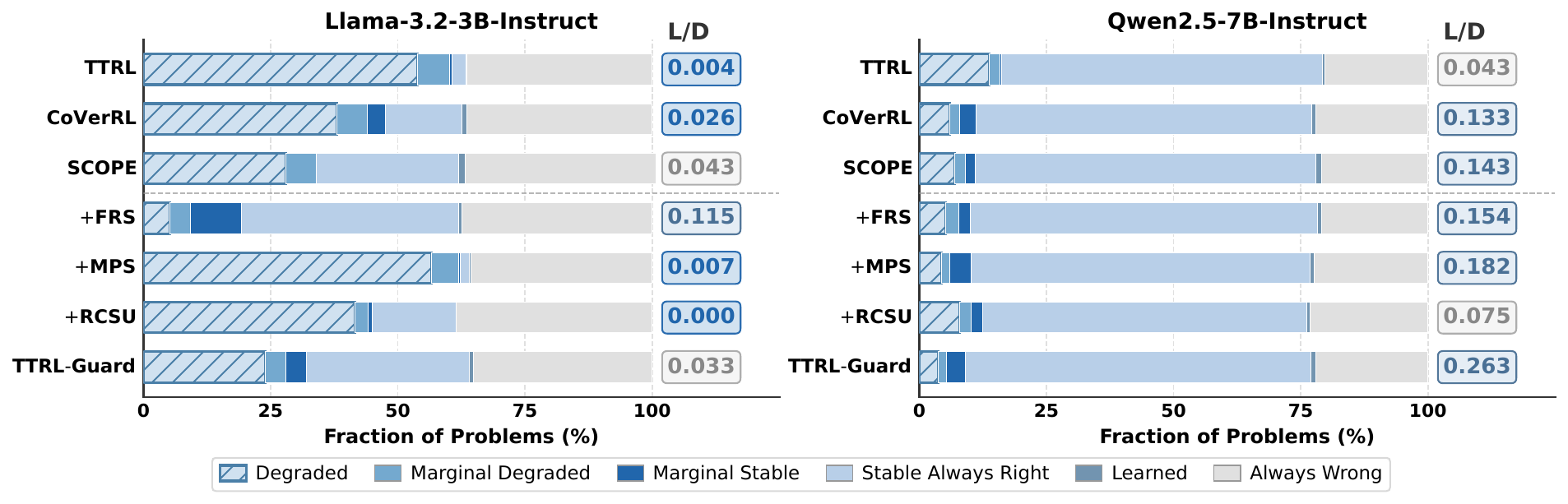}
  \caption{%
Per-problem distribution at training end on MATH-500.
Right-margin L/D values (Learned\,/\,Degraded) measure knowledge acquisition
efficiency. 
  }
  \label{fig:fate}
\end{figure*}
\begin{figure*}[t]
  \centering
  \includegraphics[width=0.8\textwidth]{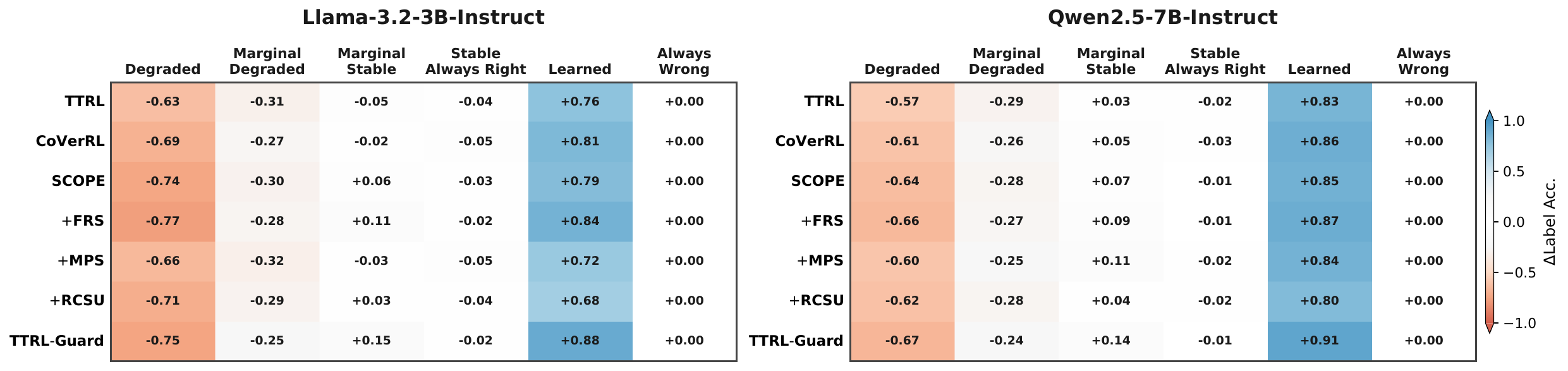}
  \caption{Mean $\Delta$LA per problem category and method on MATH-500
    (red~=~degradation, blue~=~improvement).
  }
  \label{fig:heatmap}
\end{figure*}
\paragraph{Baselines and Experimental Details.}
We compare TTRL~\cite{Zuo2025TTRL}, our proposed TTRL-Guard, and
two contemporaneous methods, SCOPE~\cite{wang2025beyond} and
CoVerRL~\cite{Pan2026CoVerRL}, which address majority-vote failures through
complementary mechanisms.  
Following TTRL, all runs sample 64 responses
per problem for majority voting and downsample 32 for policy updates. 
Detailed hyperparameter configurations are provided in Appendix~\ref{app:hyperparams}.

\paragraph{Metrics.}
We report pass@1 calculated over 4 responses per question as the primary
accuracy metric~\cite{chen2021evaluating}.
We also report the Learned/Degraded ratio (L/D), defined
as the fraction of problems whose per-problem label accuracy improves during
training divided by the fraction that declines, as a diagnostic metric
capturing the efficiency of knowledge acquisition relative to knowledge
corruption that aggregate accuracy curves conceal.

\subsection{Main Results}
\label{sec:main}

Table~\ref{tab:main} reports pass@1 for all methods across three models and four benchmarks. TTRL-Guard achieves the highest average on Qwen2.5-7B-Instruct (45.5\%) and Qwen3-4B (59.7\%), and is competitive on Llama-3.2-3B-Instruct, trailing SCOPE by only 0.4\% on average without relying on any external confidence estimator.

\paragraph{Model-capability-dependent improvements.} On stronger models (Qwen3-4B and Qwen2.5-7B-Instruct), TTRL-Guard achieves consistent gains: 59.7\% average on Qwen3-4B (leading on three of four benchmarks) and +54\% relative improvement on AIME 2025 for Qwen2.5-7B-Instruct, outperforming single-component baselines by 0.9\% on average. On weaker models (Llama-3.2-3B-Instruct), TTRL-Guard produces gains even in extreme regimes: reaching 3.3\% on AIME 2025 (where TTRL yields 0\%), and remaining competitive on MATH-500 and AMC (57.3\% and 36.8\%) compared to external-signal methods (61.7\% and 38.7\%), all without relying on external confidence estimators.

% \paragraph{Strong models benefit consistently.} On Qwen3-4B, TTRL-Guard leads on three of four benchmarks and achieves the highest average of 59.7\%, with consistent gains over TTRL across all datasets. On Qwen2.5-7B-Instruct, TTRL-Guard improves over TTRL by 54 \% on AIME 2025 and outperforms the next best single-component method, TTRL (\textit{w/} FRS), by 0.9\% on average. 

% \paragraph{Weak models on hard benchmarks.} For Llama-3.2-3B-Instruct on AIME 2025, where initial pass@1 is 0\% and TTRL yields no improvement, TTRL-Guard reaches 3.3\%, the only method to produce consistent gains in this extreme low-capacity regime. On MATH-500 and AMC, SCOPE achieves the highest single numbers (61.7\% and 38.7\%), but TTRL-Guard remains competitive (57.3\% and 36.8\%) without an external confidence estimator.

\paragraph{Ablation study.}
Each component contributes in a model-dependent manner.
On Llama-3.2-3B-Instruct, FRS is the dominant single component, raising the average to 27.1\% with the only consistent gains on AIME 2025, a result not matched by MPS or RCSU in isolation.
MPS provides the largest lift on AIME 2024, while RCSU yields modest but broad improvements across benchmarks.
On Qwen2.5-7B-Instruct, all three components are effective: FRS leads on AIME 2025 and ties for best on MATH-500; MPS achieves the top MATH-500 result; RCSU performs best on AIME 2024.
On Qwen3-4B, where TTRL already performs strongly, each single component improves over TTRL, and the full TTRL-Guard combination matches or exceeds all three ablations, confirming that FRS, MPS, and RCSU are complementary rather than redundant.

\subsection{Per-Problem Analysis}
\label{sec:fate}

\paragraph{Pass@1 masks a harm-to-benefit asymmetry that only TTRL-Guard resolves.}
Figure~\ref{fig:fate} reveals what aggregate curves conceal.
On Llama-3.2-3B-Instruct, TTRL leaves over $60\%$ of problems Degraded while genuinely learning fewer than $1\%$, yielding $\text{L/D}=0.004$, a severe asymmetry invisible in validation pass@1.
Adding FRS alone cuts the degraded fraction sharply and raises $\text{L/D}$ to $0.12$, confirming that label protection is the dominant lever, yet pass@1 rises only modestly: shielding without a fresh learning signal cannot unlock further gains.
TTRL-Guard combines FRS with MPS and RCSU to supply that signal, achieving the best pass@1 among all TTRL variants with $\text{L/D}=0.03$ and the lowest degraded fraction.
SCOPE's higher accuracy traces entirely to sharpening already-solvable problems, not to protecting degraded ones; it achieves a higher score by widening the harm gap rather than closing it.
On Qwen2.5-7B-Instruct, TTRL-Guard achieves the lowest degraded fraction across all methods, beating both SCOPE and CoVerRL, confirming that all three mechanisms act synergistically once the base model has sufficient initial capacity.

\paragraph{Accuracy degradation is concentrated and severe within the Degraded group.}
Figure~\ref{fig:heatmap} measures the depth of accuracy shift within each problem category. On Llama-3.2-3B-Instruct under standard TTRL, problems in the Degraded group experience a mean $\Delta\mathrm{LA}$ of $-0.63$: originally solvable problems driven close to zero by false-consensus reinforcement, representing irreversible capability regression. Marginal Stable problems show near-zero shifts, indicating training fails to consolidate them. TTRL-Guard is the only configuration yielding positive shifts on both models, suggesting MPS extends the correct-minority survival window long enough for gradient updates to take effect. On Qwen2.5-7B-Instruct, the Degraded group is smaller with accordingly attenuated $\Delta\mathrm{LA}$ values, consistent with fewer high-risk problems at initialization.

\begin{figure}[t]
  \centering
  \includegraphics[width=\columnwidth]{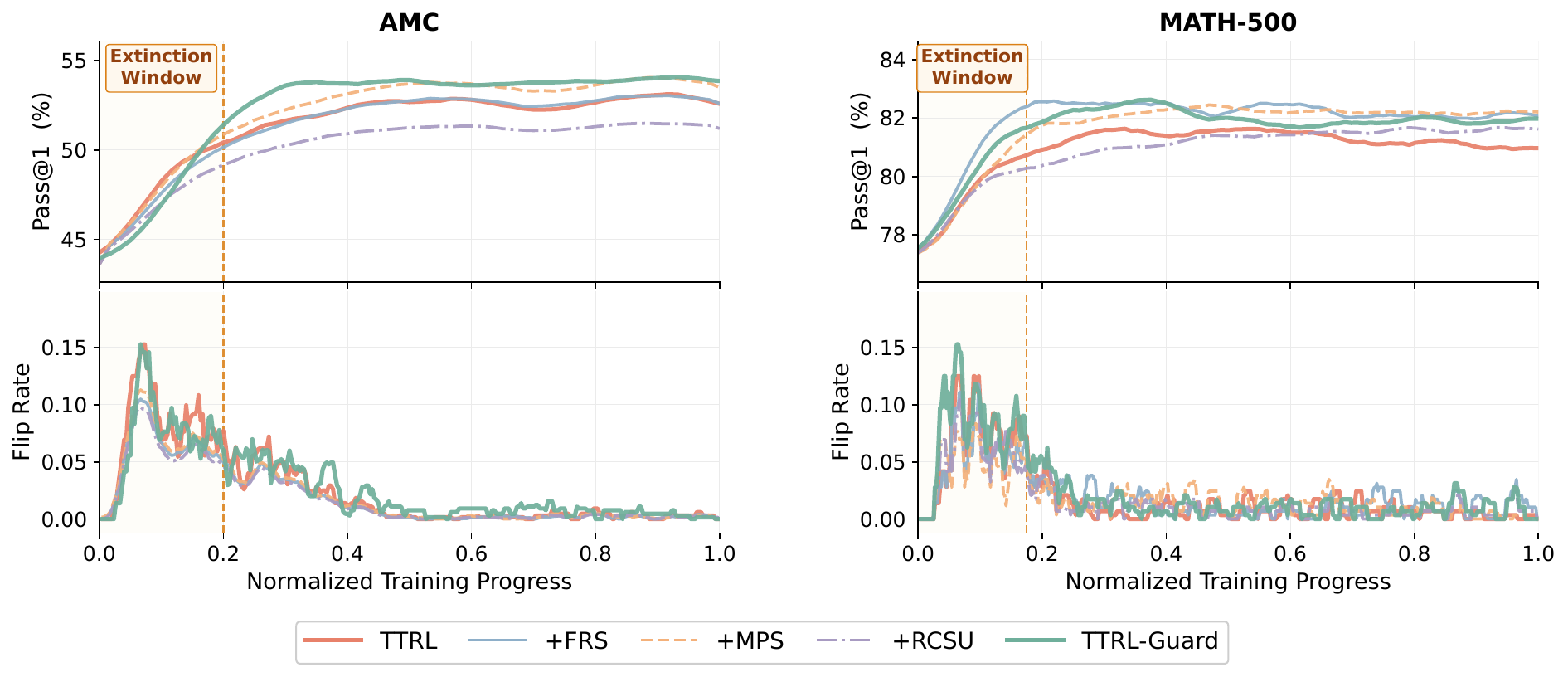}
  \caption{Training dynamics of TTRL-Guard variants on AMC and MATH-500 datasets for Qwen2.5-7B-Instruct.
  }
  \label{fig:dynamics}
\end{figure}
\paragraph{Training dynamics reveal the Extinction Window.}
Figure~\ref{fig:dynamics} exposes a striking dissociation: all five
methods converge to near-identical final Pass@1 values, yet their
Flip Rate trajectories diverge sharply during the Extinction
Window. Guard-equipped methods suppress the FR peak earlier and more
aggressively, and their accuracy advantage materialises precisely
after this window closes (e.g., TTRL-Guard reaches 53.8\%
vs.\ TTRL's 52.4\% on AMC). This temporal ordering provides causal
evidence that controlling answer-flip dynamics during the Extinction
Window is the operative mechanism behind TTRL-Guard's improvements.

\section{Related Work}
\paragraph{Unsupervised RLVR.}
Supervised RLVR has proven effective for reasoning alignment with exact feedback~\citep{guo2025deepseek}, but its reliance on ground-truth labels limits broader applicability. Unsupervised RLVR addresses this with self-generated proxy rewards, either from single-model signals such as self-certainty and entropy~\citep{zhao2025learning,agarwal2026unreasonable}, or from ensemble agreement via majority voting and semantic consistency~\citep{Zuo2025TTRL,zhang2026right,zhou2025evolving}. A complementary direction grounds rewards externally through unlabeled corpora or generation-verification asymmetries~\citep{dong2025reinforcement,simonds2025ladder}. A fundamental limitation of intrinsic approaches is that majority-vote signals can actively reinforce incorrect behaviors when consensus is corrupted~\citep{zhang2025no}, a failure our work identifies as the Correct-Answer Extinction Window.

\paragraph{Test-Time Training.}
Test-time training \cite{zhang2025test} adapts model parameters at inference time to handle distribution shifts~\citep{sun2024learning}, recently extended to LLMs~\citep{hardt2024test}. Test-Time Reinforcement Learning~\citep{Zuo2025TTRL} advances this by using majority-vote pseudo-labels to enable self-improvement without human annotation. Subsequent work ~\citep{tan2026meta} addresses majority-vote failures through external tool verification~\citep{liao2026tool}, confidence weighting~\citep{wang2025beyond}, difficulty-aware curricula~\citep{moradi2026disctt,yang2026ttcs}, and generator-verifier co-evolution~\citep{Pan2026CoVerRL}. Despite their diversity, these methods treat pseudo-label failure as a static property, leaving the dynamic degradation process during training unexamined.

\section{Conclusion}
TTRL's accuracy gains are systematically misinterpreted: most reflect sharpening of already-solvable problems, while corrupted problems far outnumber genuinely learned ones, a disparity invisible in aggregate curves. Per-problem trajectory analysis identifies the \textit{Correct-Answer Extinction Window} as the underlying mechanism, with FR as a reliable leading indicator. TTRL-Guard intervenes within this window via three complementary mechanisms, substantially reducing degradation and achieving state-of-the-art performance among label-free TTRL variants, motivating future research toward training strategies robust to untrustworthy pseudo-labels.

\section*{Limitations}

TTRL-Guard's effectiveness depends critically on the presence of correct-minority signals during the early training phase, which constrains its applicability to three regimes:

\paragraph{Low initial capability.}
When base model accuracy is near zero, the majority vote provides insufficient signal for minority-correct answers to exist. The FR signal becomes unreliable due to low answer diversity, limiting TTRL-Guard's ability to identify the extinction window.

\paragraph{High initial capability.}
When most problems are already solvable at initialization, the extinction window barely manifests, and problems rarely experience the competition phase. FRS and MPS have minimal effect, with improvements coming primarily from RCSU's selective skipping rather than extinction-window protection.

\paragraph{Hyperparameter sensitivity.}
The extinction window's characteristics (peak FR timing, decay rate) vary significantly with problem difficulty. Hyperparameters optimized on one dataset may degrade performance on others, necessitating dataset-specific tuning.

TTRL-Guard is most effective in the moderate capability-difficulty regime (approximately 30-70\% initial pass@1), where the extinction window is both present and sufficiently non-trivial for intervention.

% \section*{Limitations}

% This document does not cover the content requirements for ACL or any
% other specific venue.  Check the author instructions for
% information on
% maximum page lengths, the required ``Limitations'' section,
% and so on.

% \section*{Acknowledgments}

% This document has been adapted
% by Steven Bethard, Ryan Cotterell and Rui Yan
% from the instructions for earlier ACL and NAACL proceedings, including those for
% ACL 2019 by Douwe Kiela and Ivan Vuli\'{c},
% NAACL 2019 by Stephanie Lukin and Alla Roskovskaya,
% ACL 2018 by Shay Cohen, Kevin Gimpel, and Wei Lu,
% NAACL 2018 by Margaret Mitchell and Stephanie Lukin,
% Bib\TeX{} suggestions for (NA)ACL 2017/2018 from Jason Eisner,
% ACL 2017 by Dan Gildea and Min-Yen Kan,
% NAACL 2017 by Margaret Mitchell,
% ACL 2012 by Maggie Li and Michael White,
% ACL 2010 by Jing-Shin Chang and Philipp Koehn,
% ACL 2008 by Johanna D. Moore, Simone Teufel, James Allan, and Sadaoki Furui,
% ACL 2005 by Hwee Tou Ng and Kemal Oflazer,
% ACL 2002 by Eugene Charniak and Dekang Lin,
% and earlier ACL and EACL formats written by several people, including
% John Chen, Henry S. Thompson and Donald Walker.
% Additional elements were taken from the formatting instructions of the \emph{International Joint Conference on Artificial Intelligence} and the \emph{Conference on Computer Vision and Pattern Recognition}.


\bibliography{custom}
\appendix

\section{Problem Category Definitions}
\label{app:problem_categories}
To systematically analyze label dynamics during test-time reinforcement learning, we classify each problem according to how its pseudo-label accuracy evolves over training. We define the Initial Label Accuracy (ILA) as the mean label accuracy over the first three checkpoints and the Final Label Accuracy (FLA) as the mean over the last five. Comparing these two quantities partitions all problems into six mutually exclusive categories, summarized in Table~\ref{tab:problem_categories}.

The six categories reflect qualitatively distinct behavioral trajectories. Stable Always Right (SAR) problems are reliably solved from the outset and remain so throughout training. At the opposite extreme, Always Wrong (AW) problems are never mastered. Learned (LN) problems capture genuine knowledge acquisition, where the model initially fails but eventually succeeds. Degraded (DG) problems represent the reverse, where initially correct responses deteriorate as training progresses. The two marginal categories cover problems with intermediate initial accuracy: Marginal Degraded (MDG) exhibits a moderate decline, while Marginal Stable (MS) shows no significant change in either direction.

\begin{table*}[htbp]
\centering
\small

\begin{tabular}{@{}lll@{}}
\toprule
\textbf{Category} & \textbf{Condition} & \textbf{Description} \\
\midrule
Stable Always Right & ILA $\geq$ 0.7 and FLA $\geq$ 0.6 & Stable convergence with consistently high accuracy \\
Degraded  & ILA $\geq$ 0.5 and FLA $<$ ILA $-$ 0.2 & Initially correct but accuracy drops significantly \\
Learned   & ILA $<$ 0.15 and FLA $\geq$ 0.5 & Initially unsolvable but mastered by end of training \\
Marginal Degraded & 0.15 $\leq$ ILA $<$ 0.7 and FLA $<$ ILA $-$ 0.15 & Mid-range accuracy with notable decline \\
Marginal Stable   & 0.15 $\leq$ ILA $<$ 0.7 and FLA $\geq$ ILA $-$ 0.15 & Mid-range accuracy with no significant change \\
Always Wrong    & ILA $<$ 0.15 and FLA $<$ 0.5 & Consistently unsolvable throughout training \\
\bottomrule
\end{tabular}%
\caption{Problem categorization based on label accuracy dynamics.}
\label{tab:problem_categories}
\end{table*}

\section{Implementation Details}
\label{app:hyperparams}

Table~\ref{tab:hyperparams} summarizes the hyperparameters shared across all models, datasets, and TTRL-Guard variants. Models were trained for 60 epochs on AIME 2024/2025 and 30 epochs on AMC and MATH-500. All reported metrics represent the average of 5 independent runs.

The \textit{extinction window} is not a hyperparameter but an empirical phenomenon: a brief early-training phase in which the correct answer still holds a competitive minority before majority voting locks irreversibly onto the wrong answer. TTRL-Guard does not require its boundaries to be specified; the flip-rate signal $\mathrm{FR}_i^t$ serves as a real-time proxy, activating FRS and MPS while the window is open and handing off to RCSU once it closes.

\begin{table}[!t]
\centering
\small
\begin{tabular}{@{}lc@{}}
\toprule
\textbf{Hyperparameter} & \textbf{Value} \\
\midrule
\multicolumn{2}{l}{\textit{Training}} \\
Train batch size        & 8 \\
PPO mini-batch size     & 1 \\
PPO micro-batch size    & 2 \\
learning rate     & $5\times10^{-7}$ \\
LR warmup ratio         & 0.03 (cosine) \\
KL coefficient          & 0.00 \\
\midrule
\multicolumn{2}{l}{\textit{Rollout}} \\
Samples per prompt $K$  & 32 \\
Votes per prompt        & 64 \\
Max prompt length       & 1{,}024 tokens \\
Max response length     & 3{,}072 tokens \\
Training temperature    & 0.6 \\
Validation samples $N$  & 4 \\
\midrule
\multicolumn{2}{l}{\textit{FRS}} \\
$\lambda_1$             & 0.5 \\
$\lambda_2$             & 0.3 \\
$\tau_\mathrm{FR}$      & 0.3 \\
$\tau_\mathrm{MR}$      & 0.6 \\
Min.\ weight $w_{\min}$ & 0.1 \\
\midrule
\multicolumn{2}{l}{\textit{MPS}} \\
$\beta_\mathrm{max}$    & 0.3 \\
Minority threshold      & $\lfloor K/4 \rfloor = 8$ \\
Steady-exit steps $T_\mathrm{steady}$ & 3 \\
\midrule
\multicolumn{2}{l}{\textit{RCSU}} \\
History window $W$      & 5 \\
$\theta_\mathrm{MR}$    & 0.5 \\
Skip probability $p_\mathrm{skip}$ & 0.7 \\
Max skip per step       & 25\% of $N$ \\
\bottomrule
\end{tabular}
\caption{Hyperparameters used in all TTRL-Guard experiments.}
\label{tab:hyperparams}
\end{table}

\section{Parameter Sensitivity}

\label{app:ablation}

\begin{figure*}[htbp]
  \centering
  \begin{subfigure}[b]{0.3\textwidth}
    \centering
    \includegraphics[width=\textwidth]{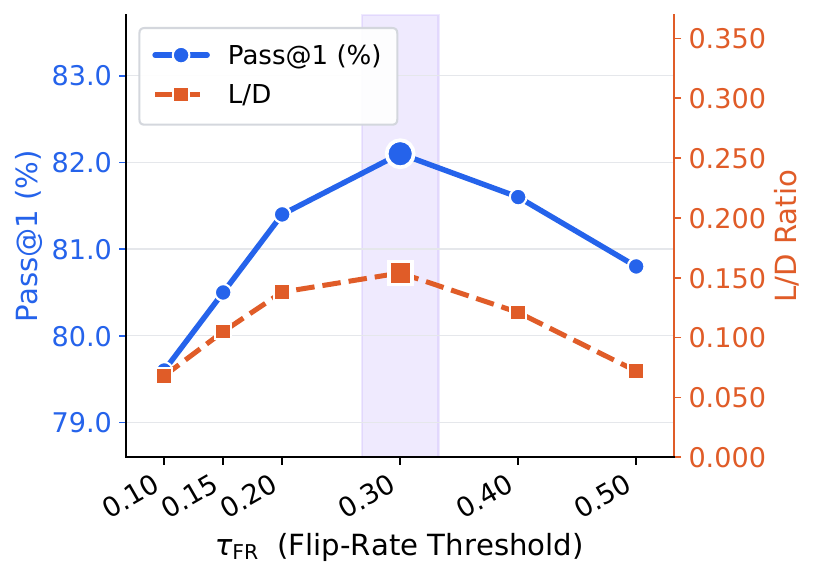}
    \caption{Sensitivity to flip-rate threshold $\tau_{\mathrm{FR}}$.}
    \label{fig:ablation_frs}
  \end{subfigure}
  \hfill
  \begin{subfigure}[b]{0.3\textwidth}
    \centering
    \includegraphics[width=\textwidth]{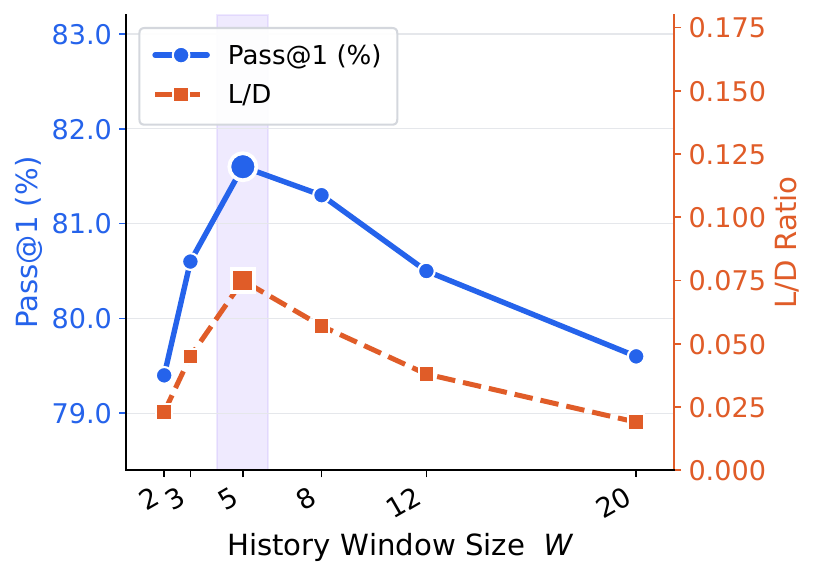}
    \caption{Sensitivity to history window size $W$.}
    \label{fig:ablation_rcsu}
  \end{subfigure}
  \hfill
  \begin{subfigure}[b]{0.3\textwidth}
    \centering
    \includegraphics[width=\textwidth]{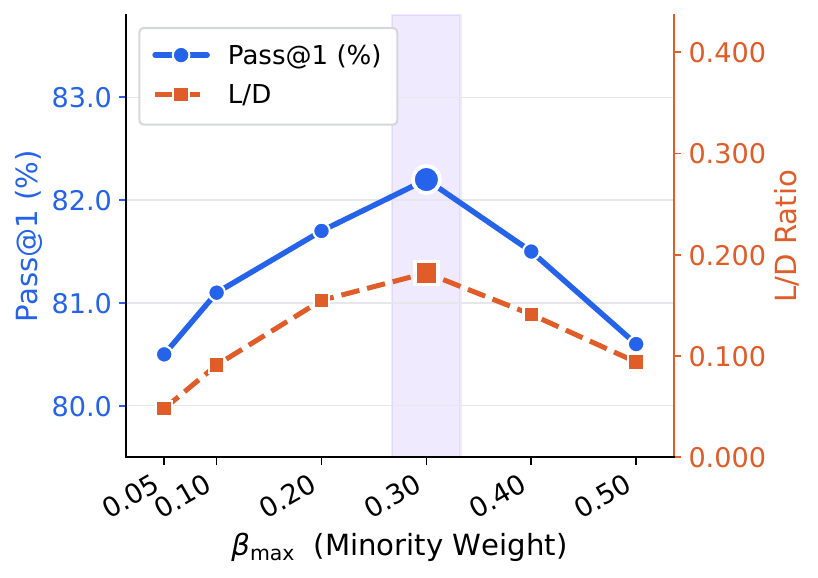}
    \caption{Sensitivity to minority weight $\beta_{\max}$.}
    \label{fig:ablation_mps}
  \end{subfigure}
  \caption{Hyperparameter sensitivity for each TTRL-Guard component (Qwen2.5-7B-Instruct / MATH-500).
           Shaded band marks the default value and larger marker highlights the optimal point.}
  \label{fig:ablation}
\end{figure*}

Figure~\ref{fig:ablation} reports the sensitivity of each component to its primary hyperparameter,
evaluated uniformly on Qwen2.5-7B-Instruct / MATH-500.
Both Pass@1 and the L/D ratio (Learned\% / Degraded\%) are shown;
the shaded band marks the recommended default.

\paragraph{FRS: flip-rate threshold $\tau_\mathrm{FR}$.}
Figure~\ref{fig:ablation_frs} sweeps $\tau_\mathrm{FR} \in [0.10, 0.50]$.
Both metrics peak at the default $\tau_\mathrm{FR}{=}0.30$, with a total swing of $2.5$ pp in Pass@1.
Small values ($\tau_\mathrm{FR} \leq 0.15$) over-trigger FRS, suppressing correct-minority gradients alongside noisy ones and degrading both Pass@1 and L/D.
Large values ($\tau_\mathrm{FR} \geq 0.40$) prevent FRS from firing on mildly unstable problems,
allowing wrong-answer lock-in to proceed unchecked and collapsing L/D sharply.
The method is robust for $\tau_\mathrm{FR} \in [0.20, 0.40]$.

\paragraph{RCSU: history window $W$.}
Figure~\ref{fig:ablation_rcsu} sweeps $W \in \{2, 3, 5, 8, 12, 20\}$.
The default $W{=}5$ achieves the best Pass@1 ($81.6\%$) and L/D ($0.075$).
Short windows ($W \leq 3$) produce noisy match-rate estimates, causing HighRisk flags to fire on transient spikes and incorrectly skipping learnable problems, which suppresses Learned\% and hurts accuracy.
Large windows ($W \geq 12$) detect reward lock-in too slowly; wrong-answer gradients accumulate before skipping activates, driving Degraded\% up and Pass@1 down.
L/D is more sensitive to $W$ than Pass@1, confirming that RCSU primarily guards training dynamics quality.

\paragraph{MPS: maximum minority weight $\beta_\mathrm{max}$.}
Figure~\ref{fig:ablation_mps} sweeps $\beta_\mathrm{max} \in [0.05, 0.50]$.
At $\beta_\mathrm{max}{=}0.05$, the minority reward is negligible and results approach the plain TTRL baseline ($80.5\%$);
L/D remains near its unguarded level, confirming that minority-correct gradients are not being preserved.
Both metrics peak at the default $\beta_\mathrm{max}{=}0.30$.
Above $0.30$, excessive upweighting amplifies incorrect minority answers (any response with $\geq\lfloor K/4 \rfloor$ votes qualifies for $\mathcal{M}_i^t$, not only correct ones),
introducing gradient noise that degrades both Pass@1 and L/D.

\section{Computational Overhead Analysis}

\label{app:overhead}

A critical concern when introducing per-problem state tracking is the potential increase in training latency. We report the relative change in wall-clock time per training step compared to TTRL (positive values indicate slowdown; negative values indicate speedup) in Table~\ref{tab:overhead}. FRS incurs modest overhead from per-problem state monitoring, and MPS adds a small additional loss branch; both introduce only positive overhead regardless of model size. RCSU, by contrast, skips gradient updates for high-risk problems, reducing the number of tokens processed during the backward pass. Its speedup grows with model size, as the backward pass becomes increasingly dominant for larger models. On Llama-3.2-3B-Instruct, the skip savings are limited, so the TTRL-Guard pipeline incurs a small net overhead of +3.8\%. On Qwen2.5-7B-Instruct, RCSU's skipping outweighs the overhead of FRS and MPS, yielding a net speedup of 29.3\% for the full pipeline.

\begin{table}[htbp]
\centering
\small
\renewcommand{\arraystretch}{0.82}
\begin{tabular}{@{}l cc@{}}
\toprule
\textbf{Method} & \textbf{Llama-3.2-3B} & \textbf{Qwen2.5-7B} \\
\midrule
TTRL              & ---       & ---       \\
~~+ FRS           & +5.3\%    & +5.0\%    \\
~~+ MPS           & +13.4\%   & +13.2\%   \\
~~+ RCSU          & $-$6.0\%  & $-$34.9\% \\
TTRL-Guard        & +3.8\%    & $-$29.3\% \\
\bottomrule
\end{tabular}
\caption{Relative change in wall-clock time per step vs.\ TTRL on MATH-500. Negative values indicate speedup.}
\label{tab:overhead}
\end{table}

\begin{table*}[!t]
\centering
\small
\setlength{\tabcolsep}{5pt}
\renewcommand{\arraystretch}{0.82}

\begin{tabular}{@{}ll rrrrrr@{}}
\toprule
\textbf{Dataset} & \textbf{Method} & \textbf{Degraded} & \textbf{Marg. Deg.} & \textbf{Total Deg.} & \textbf{Learned} & \textbf{Stable AR} & \textbf{Always Wrong} \\
\midrule
\multicolumn{8}{c}{\textit{\textbf{Llama-3.2-3B-Instruct}}} \\
\midrule
\multirow{2}{*}{MATH-500}
& TTRL            & 53.8\% &  6.4\% & \textbf{60.2\%} & 0.2\% &  2.8\% & 36.4\% \\
& TTRL-Guard      & 24.0\% &  4.0\% & \textbf{28.0\%} & 0.8\% & 32.0\% & 35.2\% \\
\cmidrule(l){2-8}
\multirow{2}{*}{AMC}
& TTRL            & 12.0\% &  3.6\% & \textbf{15.6\%} & 1.2\% & 24.1\% & 55.4\% \\
& TTRL-Guard      &  5.0\% &  3.0\% & \textbf{8.0\%}  & 2.4\% & 28.3\% & 55.4\% \\
\cmidrule(l){2-8}
\multirow{2}{*}{AIME 2024}
& TTRL            &  3.3\% &  3.3\% & \textbf{6.6\%}  & 3.3\% & 13.3\% & 76.7\% \\
& TTRL-Guard      &  0.0\% &  3.3\% & \textbf{3.3\%}  & 6.7\% & 16.7\% & 73.3\% \\
\midrule
\multicolumn{8}{c}{\textit{\textbf{Qwen2.5-7B-Instruct}}} \\
\midrule
\multirow{2}{*}{MATH-500}
& TTRL            & 13.8\% &  2.0\% & \textbf{15.8\%} & 0.6\% & 63.0\% & 20.2\% \\
& TTRL-Guard      &  3.8\% &  1.6\% & \textbf{5.4\%}  & 1.0\% & 68.0\% & 22.0\% \\
\cmidrule(l){2-8}
\multirow{2}{*}{AMC}
& TTRL            &  1.2\% &  1.2\% & \textbf{2.4\%}  & 0.0\% & 48.2\% & 47.0\% \\
& TTRL-Guard      &  0.0\% &  1.2\% & \textbf{1.2\%}  & 1.2\% & 50.6\% & 44.6\% \\
\cmidrule(l){2-8}
\multirow{2}{*}{AIME 2025}
& TTRL            &  3.3\% &  0.0\% & \textbf{3.3\%}  & 0.0\% & 13.3\% & 76.7\% \\
& TTRL-Guard      &  0.0\% &  3.3\% & \textbf{3.3\%}  & 3.3\% & 20.0\% & 73.3\% \\
\bottomrule
\end{tabular}
\caption{Per-problem fate distributions for Llama-3.2-3B-Instruct and Qwen2.5-7B-Instruct across four datasets. TotalDeg is the sum of Degraded and Marginal Degraded.}
\label{tab:full_fates}
\end{table*}

% \section{Visualizing the "Illusion of Learning"}
% \label{app:illusion_learning}

% In the main text, we argue that the headline accuracy gains of TTRL mostly reflect the consolidation of existing knowledge rather than the acquisition of new capabilities. Figure~\ref{fig:illusion_learning} visually corroborates this claim by plotting the migration of problems in the Label Accuracy (LA) vs. Match Rate (MR) spectrum.

% Notice that the problem distribution does not shift uniformly from the lower-left (unsolved) to the upper-right (solved). Instead, problems that were already partially solvable (middle spectrum) are forcefully pushed to the extreme right (MR $\approx 1.0$), constituting a "sharpening" effect. Meanwhile, the dense cluster of always-wrong problems remains largely stagnant. This visualization exposes the fragility of unsupervised TTRL: the mechanism inherently struggles to bootstrap zero-to-one reasoning without external guidance.

% \begin{figure}[htbp]
%   \centering
%   \includegraphics[width=\columnwidth]{figures/fig1_spectrum_la_mr.pdf}
%   \caption{Visualizing the "Illusion of Learning": The migration of problem distributions across the Label Accuracy (LA) and Match Rate (MR) spectrum before and after TTRL training. The visible shift primarily demonstrates the polarization of already-known concepts rather than genuine learning.}
%   \label{fig:illusion_learning}
% \end{figure}

\section{The Capability-Difficulty Mismatch}

\label{app:capability_mismatch}

To demonstrate the consistency of asymmetric degradation and the effectiveness of TTRL-Guard across varying task difficulties, Table~\ref{tab:full_fates} reports the complete per-problem fate distributions for Llama-3.2-3B-Instruct and Qwen2.5-7B-Instruct across AIME 2024, AIME 2025, AMC, and MATH-500.

The results reveal a consistent structural pattern: degradation severity is strongly correlated with the capability-difficulty mismatch between the model and the dataset. When a dataset is entirely beyond a model's capability (e.g., Llama-3.2-3B-Instruct on AIME 2025, where almost all problems fall into Always Wrong), majority voting rarely produces a false consensus, so TotalDeg remains near zero and TTRL-Guard has little room to intervene. At the other extreme, when a strong model faces an easy dataset (e.g., Qwen2.5-7B-Instruct on AMC), most problems are already stably correct, leaving only a small fraction susceptible to degradation. The most severe degradation occurs in the moderate-mismatch regime, exemplified by Llama-3.2-3B-Instruct on MATH-500, where TotalDeg reaches 60.2\% under TTRL. In this regime, the model is capable enough to generate plausible but incorrect answers that attract a majority, yet not capable enough to recover once the wrong pseudo-label locks in. TTRL-Guard reduces TotalDeg by 32.2 percentage points in this setting, from 60.2\% to 28.0\%, while simultaneously increasing Stable AR from 2.8\% to 32.0\%, confirming that the interventions protect knowledge rather than merely suppressing updates.

\begin{figure*}[!t]
  \centering

  \begin{subfigure}[t]{0.24\textwidth}
    \centering
    \includegraphics[width=\linewidth]{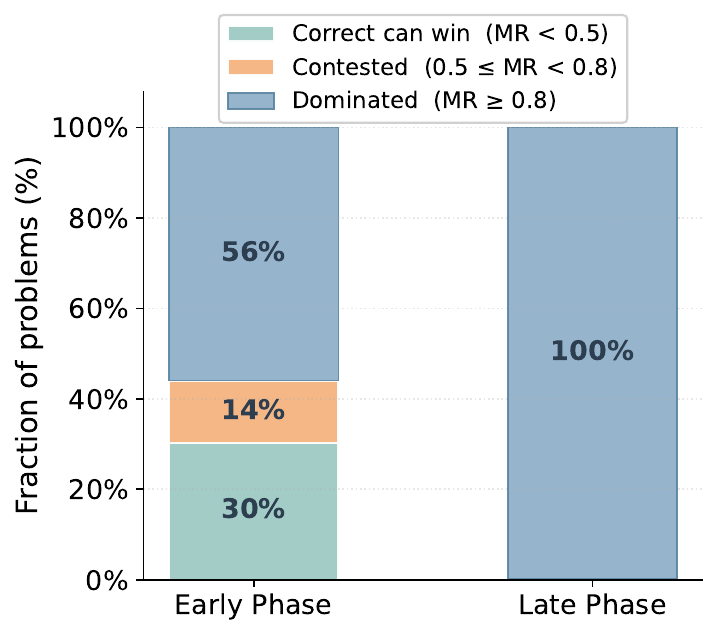}
    \caption{Wrong-answer MR distribution, early vs.\ late training.}
    \label{fig:minority_elim_a}
  \end{subfigure}
  \begin{subfigure}[t]{0.24\textwidth}
    \centering
    \includegraphics[width=\linewidth]{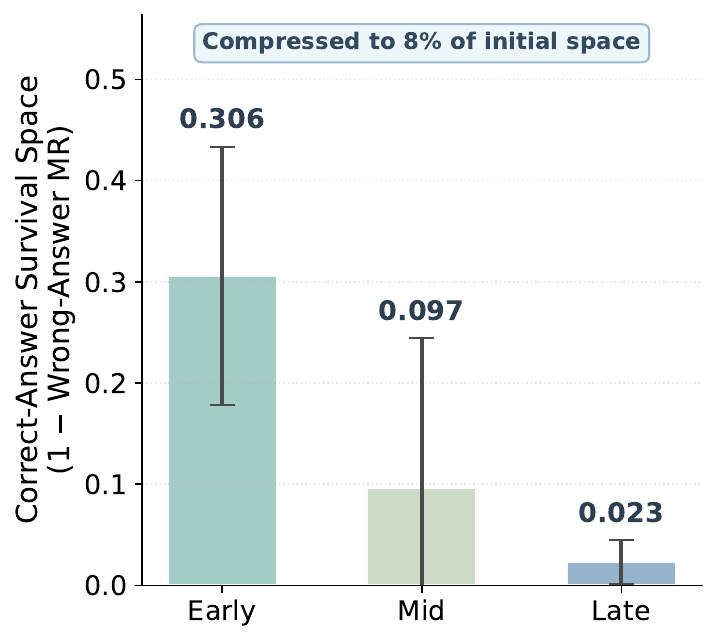}
    \caption{Correct-answer survival space across training phases.}
    \label{fig:minority_elim_b}
  \end{subfigure}
  \begin{subfigure}[t]{0.25\textwidth}
    \centering
    \includegraphics[width=\linewidth]{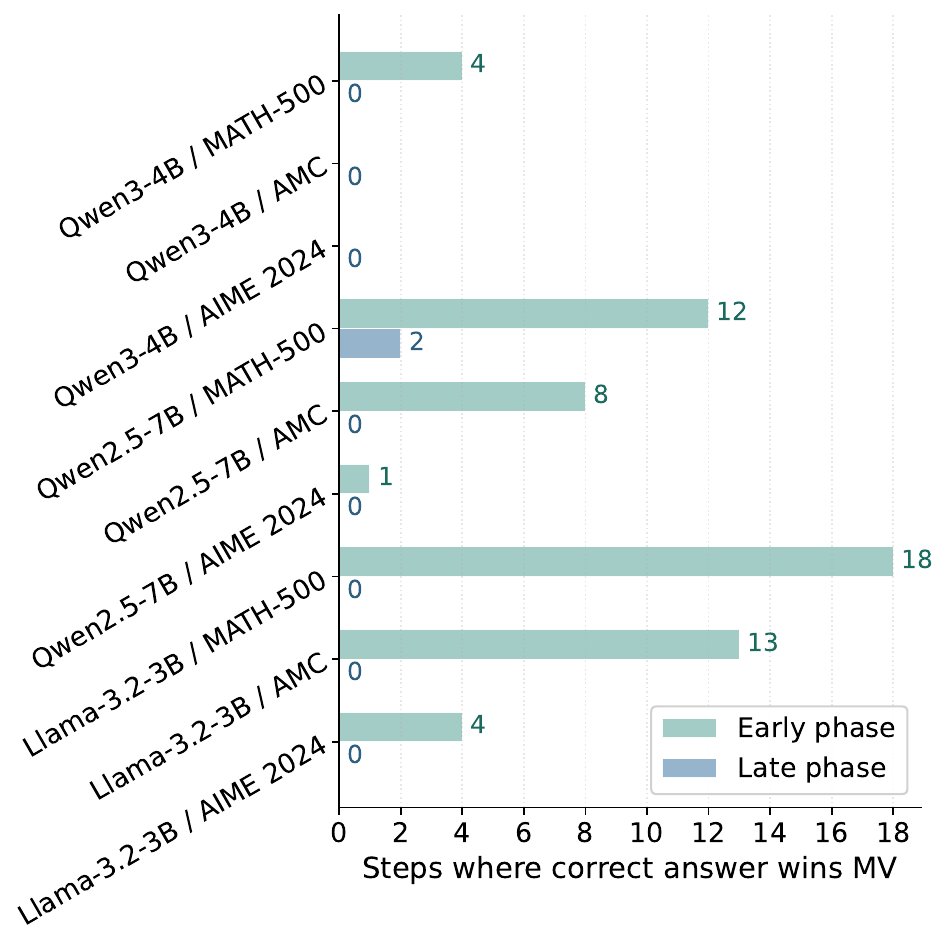}
    \caption{Steps where the correct answer wins the majority vote.}
    \label{fig:minority_elim_c}
  \end{subfigure}
  \begin{subfigure}[t]{0.25\textwidth}
    \centering
    \includegraphics[width=\linewidth]{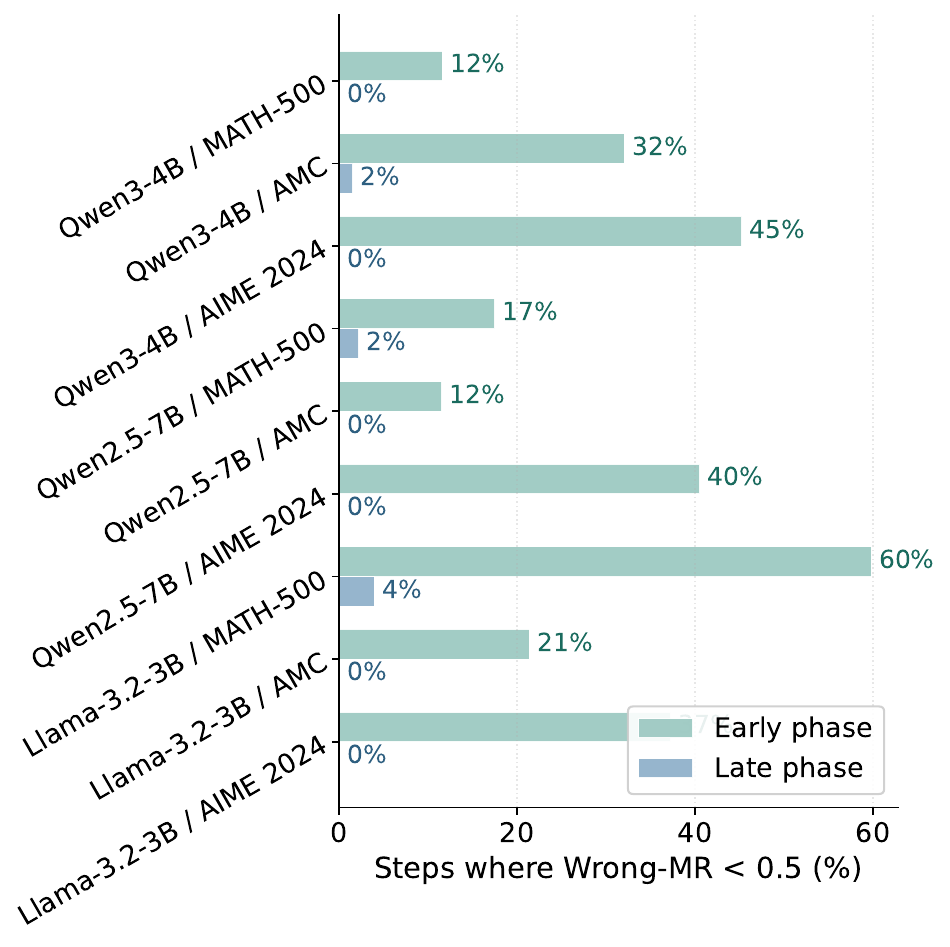}
    \caption{Extinction window width per model and dataset.}
    \label{fig:minority_elim_d}
  \end{subfigure}

  \caption{The Scissor Effect underlying the Correct-Answer Extinction
    Window. (a) Even in early training, most problems are already
    dominated by the wrong answer; by late training this fraction approaches
    100\%. (b) The correct-answer survival space collapses
    monotonically across phases. (c, d) The window of
    correct-answer competitiveness is narrow and confined to the early
    phase, motivating MPS to capture these signals before vanishing.}
  \label{fig:minority_elimination}
\end{figure*}
% \begin{figure*}[!t]
%   \centering
%   \includegraphics[width=\textwidth]{figures/fig8_spectrum_fr_ts.pdf}
%   \caption{Temporal Heterogeneity of the Flip Rate across different problem difficulties. The timing of the Extinction Window varies drastically depending on the initial prompt ability, necessitating per-prompt dynamic tracking.}
%   \label{fig:fr_heterogeneity}
% \end{figure*}
% \section{Microscopic Dynamics of the Consensus Trap}
% \label{app:microscopic_dynamics}

\section{Microscopic Dynamics of the Consensus Trap}
\label{app:microscopic_dynamics}
To further illustrate the mechanics of Asymmetric Degradation, we isolate
the trajectory of problems that fall into the Degraded category
(initially correct, but corrupted by the end of training).

When tracing the step-by-step voting distribution of a typical Degraded
problem, we observe a distinct \textbf{Scissor Effect} that drives the
Extinction Window, as visualized in Figure~\ref{fig:minority_elimination}.
The dynamics unfold in three phases:

\begin{enumerate}

  \item \textbf{Phase 1 (Competition).}
    In the early training steps, the correct answer and a highly plausible
    incorrect answer co-exist, competing closely for the majority vote.
    As shown in Figure~\ref{fig:minority_elim_a}, aggregated over all three
    models (Llama-3.2-3B-Instruct, Qwen2.5-7B-Instruct, and Qwen3-4B) on
    AIME 2024, AMC, and MATH-500, roughly 30\% of Degraded problems still
    have a wrong-answer Match Rate below 0.5 in the early phase, meaning
    the correct answer retains a chance to win the majority vote.

  \item \textbf{Phase 2 (Suppression).}
    Due to the winner-takes-all nature of the standard GRPO loss, any step
    in which the incorrect answer wins the majority vote heavily penalises
    the correct answer. The incorrect answer systematically suppresses the
    correct answer's log-probabilities, producing a visible crossover.
    Figure~\ref{fig:minority_elim_b} quantifies this collapse: the
    correct-answer survival space $1 - \overline{\mathrm{MR}}$, averaged
    across all model and dataset combinations, shrinks monotonically from
    the early to the late phase, compressing to a small fraction of its
    initial value.

  \item \textbf{Phase 3 (Lock-in).}
    Once the Match Rate of the incorrect answer crosses a critical
    threshold, the flip rate collapses and the correct answer's sampling
    probability drops near zero, finalising irreversible degradation.
    Figures~\ref{fig:minority_elim_c} and~\ref{fig:minority_elim_d}
    confirm this pattern across all nine model-dataset pairs: the number of
    steps in which the correct answer wins the majority vote, and the
    fraction of steps in which the wrong-answer MR remains below 0.5, both
    drop sharply from the early phase to the late phase.
    The extinction window is therefore narrow, transient, and
    consistent across architectures and difficulty levels.

\end{enumerate}
TTRL provides no mechanism to detect or counteract this crossover. TTRL-Guard addresses this gap through MPS, which maintains a non-zero gradient for the correct answer as long as it retains any minority presence in the rollout batch. This delays the crossover and extends the window in which the model can self-correct before degradation becomes irreversible.

\end{document}